\documentclass{elsart}

\usepackage{graphicx}

\begin{document}

\begin{frontmatter}

\title{Single-solution Random 3-SAT Instances} 

\author{Marko \v Znidari\v c}
\address{Department of Quantum Physics, University of Ulm,
       D-89069 Ulm, Germany
      }

\begin{abstract}
We study a class of random 3-SAT instances having exactly one solution. The properties of this ensemble considerably differ from those of a random 3-SAT ensemble. It is numerically shown that the running time of several complete and stochastic local search algorithms monotonically increases as the clause density is decreased. Therefore, there is no easy-hard-easy pattern of hardness as for standard random 3-SAT ensemble. Furthermore, the running time for short single-solution formulas increases with the problem size much faster than for random 3-SAT formulas from the phase transition region.
\end{abstract}

\end{frontmatter}

\section{Introduction}
\label{intro}
The propositional satisfiability problem is one of the most studied problems 
in computer science. The most prominent one is the 3-satisfiability (3-SAT) 
problem. It consists of determining if there exists an assignment of truth 
values to a set of boolean variables such that a given 3-SAT formula is 
satisfied. A 3-SAT formula involving $n$ variables is a conjunction 
(logical AND) of $m$ clauses, each clause being a disjunction (logical OR) 
of $3$ literals (a literal is a variable or its negation). 3-SAT problem is 
important from a theoretical as well as from a practical point of view. 
On the theoretical side, it is a paradigmatic example of a NP-complete (NPC) 
problem. Historically, it was the first problem to be shown by 
Cook~\cite{Cook:71} to be NPC. The algorithmic complexity of 3-SAT problem 
is connected to various computational complexity issues, most notably to 
the famous ``P=NP?'' question which is one of the most important unsolved 
problems in mathematics and computer science~\cite{Clay:00}. 
On the practical side, 3-SAT solving algorithms are used in the industry. 
Because any circuit involving logical operations can be converted to a 3-SAT 
formula they can also be used for verification of 
microprocessors~\cite{Velev:01,Bjesse:01}. 3-SAT solving can also be related 
to deductive reasoning: given a set of facts (statements) $\phi$, a new 
statement $C$ can be deduced if a union $\phi \cup \{ \lnot C\}$ is not 
satisfiable, i.e. we arrive at the contradiction assuming the 
negation $\lnot C$. 

Having hard~\footnote{By hard we mean that the number of steps needed by a 
given algorithm to solve the problem is larger than for most instances of the 
same size.} 3-SAT instances at hand is important for several reasons. First, 
it might help in understanding what makes 3-SAT problems, and generally NPC 
ones, so difficult at all. Second, hard instances are also actively sought 
for algorithm testing, enabling the design of better algorithms. There are 
basically two classes of 3-SAT instances used in testing. Those coming from 
the real world applications mentioned above and artificially generated ones 
that are thought to be hard. For the later ones one usually uses the 
so-called uniform random 3-SAT ensemble. An important discovery was that 
among random 3-SAT instances hard ones are found around the phase 
transition~\cite{Selman:92a,Selman:92,Kirkpatrick:94,Crawford:93,Selman:96,Crawford:96}, 
where the average formula changes from being satisfiable to being unsatisfiable. Connected to the phase 
transition phenomenon, it is believed that for NPC problems one typically 
has a pattern of ``easy-hard-easy'' problem difficulty as some parameter is 
varied, with the peak difficulty occurring at the phase transition.

If considering single 3-SAT instances one can ask for instance if a 
given formula is harder than some other one. For NP-completeness the 
relevant question is how the running time of the hardest instance 
(for a given $n$) increases with its size. It is important to realize that 
if we want to talk about perhaps more interesting statistical 
properties (e.g. scaling of the running time, phase transition etc.) one has to 
specify an ensemble of 3-SAT instances, that is defining a measure, i.e. a probability do draw some instance. Therefore, a phase 
transition phenomenon is not an inherent property of 3-SAT problem alone but
of the measure, i.e. it is induced by the drawing procedure (for instance 
for uniform random 3-SAT). But whereas for physical systems there exists a 
natural measure, there is no such thing for mathematical problem like 3-SAT.
 Physical systems have a distinguished quantity called the energy which 
induces the canonical measure. The canonical measure depends on the temperature and as this parameter is varied a phase transition can occur. For 3-SAT problem there is no such ``natural'' measure. For the most frequently studied random 3-SAT ensemble literals occur in clauses with equal probability. While this might seem a least biased choice there is no a priori reason why such a measure is better than any other. As the random 3-SAT ensemble is just one of many possible ones it is in a way surprising than since the discovery of phase transition in 3-SAT most 
studies have been concerned with random 3-SAT ensemble 
(for earlier study of the so-called ``random clause length'' SAT ensemble 
see, e.g.~\cite{Goldberg:79}). In fact, the measure for random 3-SAT is based on the syntax of the particular encoding of the problem, therefore it is not directly related to the problem structure. It might be useful to consider measure which directly depends on an inherent 3-SAT property, e.g. on the number of satisfying assignments. An interesting question then is how do the 
properties of an ensemble depend on the chosen measure? One of the initial ideas was~\cite{Cheeseman:91} that NP-completeness is intimately connected to the phase transition 
phenomenon. Does one therefore have an easy-hard-easy pattern also for other 
3-SAT ensembles? Also, what is the hardness of instances from other ensembles, are they harder than instances from the phase transition for random 3-SAT?    

In this work we will try to answer some of these questions. We will study an 
ensemble of random 3-SAT instances having a single satisfying assignment. By 
an empirical study of the running time of several complete and incomplete algorithms we are going to show that there is no hardness-peak for this 
ensemble. In addition, such single-solution instances also seem to be much 
harder than 3-SAT instances from random 3-SAT phase transition region. We have been actually drawn to 3-SAT problems with one solution 
while studying quantum adiabatic algorithm for 3-SAT. Quantum adiabatic 
algorithm~\footnote{For the present status of quantum algorithms see, e.g. the overview~\cite{Shor:04}.} attracted attention because there has been 
numerical evidence (for small problems) that the running time for random 3-SAT 
instances from the phase transition region increases only quadratically with 
the size $n$~\cite{Farhi:01}. Contrary to that, the scaling of 
adiabatic algorithm for instances with a single solution seems to be 
rather exponential~\cite{Znidaric:05}. It is therefore also interesting to 
compare the performance of classical algorithms for these two ensembles, 
particularly because some stochastic local search algorithms seem to be 
fairly efficient on random 3-SAT problems from the phase transition. Random 
instances with a constant number of satisfying assignments could also 
improve our understanding of phase transition phenomenon in random 3-SAT. Let us first briefly review some known facts about the uniform random 3-SAT ensemble. 

\section{Uniform Random 3-SAT}

A 3-CNF formula is a logical statement involving $n$ boolean variables $b_i$. 
It consists of $m$ clauses $C_i$ in conjunction (logical AND $=\land$), 
$C_1 \land C_2 \land \cdots \land C_m$, where each clause $C_i$ is a 
disjunction (logical OR $=\lor$) of 3 literals, where a literal is a 
variable $b_i$ or its negation $\lnot b_i$ (logical NOT$=\lnot$). 
3-SAT problem is to decide whether a given 3-CNF formula, denoted by 
$\phi$, is satisfiable, i.e. whether there exists a truth assignment of 
variables $b_i$ such that $\phi$ is true. Such prescription is called a 
solution, the number of which will be denoted by $r$. A given assignment 
of all variables will be called a state. An instance from a uniform random 
3-SAT ensemble is generated by drawing three different random variables for 
each clause and negating each with probability $\frac{1}{2}$. 
The number of different clauses is therefore $s=8{n \choose 3}$. 

For random 3-SAT problems it has been established that the relevant order 
parameter for the phase transition is a ratio of the number of clauses and 
the number of variables, the so-called clause density 
$\alpha=m/n$~\cite{Selman:92a,Crawford:93,Selman:96}. The critical value 
$\alpha_{\rm c}\approx 4.25$ for the transition between satisfiability and 
unsatisfiability coincides with the peak in hardness, i.e. the peak in the running time 
of an algorithm see, e.g. Figure~\ref{fig:DPLLphase}. Below this critical 
$\alpha_{\rm c}$ random problems are satisfiable with high probability as 
they are underconstrained, while above it they are 
unsatisfiable because they are overconstrained. The width of the transition 
region in parameter $\alpha$ has been shown to decrease as $\sim 1/n^{2/3}$ 
with the number of variables~\cite{Kirkpatrick:94}. Still, we do not
have an exact expression for the location of the transition point. The 
best present proved bounds for the critical $\alpha$ are $3.42$ for 
satisfiability border~\cite{Kaporis:02} and $4.506$ for unsatisfiability 
border~\cite{Dubois:03}, therefore $3.42 < \alpha_{\rm c} < 4.506$. See 
the review by~\cite{Cook:97} for references about the location of the 
transition point.

A very fruitful approach to 3-SAT problem is to convert it to a spin glass 
system and then use various powerful statistical methods. One can convert 
a given 3-SAT formula to a (classical) Hamiltonian by the following 
simple prescription: for each variable $b_i$ a spin variable $S_i$ is 
assigned with the value $S_i=1$ corresponding to $b_i=1$ and $S_i=-1$ 
for $b_i=0$. The Hamiltonian, whose expectation value counts the number 
of unsatisfied clauses by a state, is a sum of terms for each clause $C_i$, 
$H=\sum_{i=1}^{m}{H_{C_i}}$, where the rule for the Hamiltonian $H_{C_i}$ describing the clause 
$C_i$ can be best seen from an example, 
$(b_2 \lor \lnot b_4 \lor b_5) \longrightarrow H_{C_i}=\frac{1}{8}(1-S_2)(1+S_4)(1-S_5)$, i.~e. the signs in front of spin variables are determined by clauses. A 
solution will therefore have energy $0$, and the question of satisfiability 
is translated into the question about the ground state of $H$ with energy 
zero. Statistical methods have been used to estimate $\alpha_{\rm c}$ and 
to show that the number of solutions just below the transition point is 
exponentially large~\cite{Monasson:96}, so the transition is reminiscent 
of a discontinuous (1st order) phase transition in statistical physics. Analysis of the phase space 
structure also resulted in a new survey propagation 
algorithm~\cite{Mezard:02}. The hardness of instances at the 
transition point has been connected with the discontinuous occurrence of 
a ``backbone''. A backbone is a set of variables that are fully constrained, 
i.e. have the same value in all solutions. Below $\alpha_{\rm c}$ the 
backbone is zero, while it is nonzero (and bounded away from zero) above 
$\alpha_{\rm c}$~\cite{Monasson:99}. If the backbone is large and the 
problem is overconstrained a backtracking algorithm will quickly ``realize'' 
it made a wrong assignment. On the other hand if the backbone is small and 
the problem is underconstrained there are many ``good'' beginning 
assignments which will lead to the solution.

\section{Related work}
In this section we will give a list of related studies that deal 
with the subjects covered in the present paper. This includes studies of 
instances with a fixed number of solutions, scaling of the running time 
with $n$, generating methods for hard instances and various results about 
the difficulty of short 3-SAT formulas (having small $m$). 

Most of the studies of random 3-SAT ensemble have been concerned with the 
computational cost at a constant $n$ as a function of the ratio $m/n$, 
where the characteristic phase transition-like curve is observed. 
This is in a way surprising because for 
the computational complexity (and also for practical applications) it is 
the scaling of running time as the problem size increases which is important, 
i.e. changing $n$ at fixed $m/n$. Exponential scaling with $n$ has been 
numerically observed near the critical point~\cite{Crawford:93,Crawford:96} 
for random 3-SAT as well as above it (albeit with a smaller exponent). Recently the scaling 
with $n$ has been studied and the transition from polynomial to exponential 
complexity has been observed below $\alpha_{\rm c}$~\cite{Coarfa:03}, again for random 3-SAT.

There has been numerical evidence~\cite{Selman:92a,Gent:94,Gent:96} that below $\alpha_{\rm c}$ short instances of 3-SAT as well as of graph coloring~\cite{Hogg:94} can be hard. With respect 
to the formula size
 an interesting rigorous result is~\cite{Beame:98,Beame:02} that an 
ordered DPLL algorithm needs an exponential time $2^{\Omega(n/\alpha)}$ to 
find a resolution proof of an unsatisfiable 3-SAT instance. Note that the
 coefficient of the exponential growth increases with decreasing $\alpha$, i.e. short formulas are harder. For our ensemble of single-solution formulas we 
will find the same result.

Generating methods for 3-SAT problems having one solution have been devised
 employing the Latin square problem~\cite{Achlioptas:00} as well as transforming the factorization to 3-SAT~\cite{Horie:96}. In both cases the parameter 
$\alpha$ of the resulting 3-SAT instances grows with the problem size $n$. 
One can also use the conversion of some other NPC problem to 
3-SAT~\cite{Cadoli:05}. Hard instances in the underconstrained region can 
be generated by embedding a smaller unsatisfiable subproblem~\cite{Bayardo:96} into a larger instance. Ferromagnetic phase transition in a spin glass has 
been exploited to generate hard satisfiable 3-SAT instances in the overconstrained region, $m/n > \alpha_{\rm c}$~\cite{Barthel:02}. Hard instances can also be generated by hiding satisfying assignments~\cite{Achlioptas:05}. This work has been extended~\cite{Haixia:05} to produce even harder instances, particularly for stochastic local search methods. For instance, the number of necessary Walksat steps grows exponentially as $\propto 2^{0.1n}$ for the hardest instances generated.

As regards the connection between the number of solutions and the formula 
difficulty there have been several works, but none studied in detail how the 
time scales if the number of solutions is held fixed. In~\cite{Smith:96} it has 
been found that for constraint satisfaction problem the difficulty 
monotonically increases by decreasing the phase transition parameter.
Later it was found~\cite{Mammen:97} that 
the existence of the peak in hardness can sensitively depend in the ensemble 
and the algorithm used. Hoos~\cite{Hoos:98} found a correlation between the 
number of solutions and the problem difficulty, i.e. instances with less solutions tend to be harder for stochastic local search methods, see also~\cite{Clark:96}. Problems having a small backbone seem to have stronger correlation between the number of solutions and a local search difficulty~\cite{Singer:00}.

\section{Random 3-SAT instances with one solution}
\label{r1}
Although the precise understanding of the phase transition phenomenon in 
random 3-SAT is still lacking a frequent heuristic explanation for 
its occurrence (see, e.g.~\cite{Williams:94}) is the combination of a decreasing
number of solutions as $\alpha$ is increased and at the same time increased
pruning of the search three, resulting in the maximal complexity for 
some intermediate $\alpha$. This ``explanation'' classifies 3-SAT instances
according to the number of solutions they have. Therefore, an ensemble of 
random 3-SAT instances having a constant number of solutions would 
presumably tell us also something about the phase transition itself. As a 
second motivation point to choose a single-solution ensemble is the fact that for some applications~\cite{Horie:96,Achlioptas:00} a single-solution 3-SAT instances might be more ``natural'' choice than random 3-SAT. 

\begin{figure}[ht!]
\centerline{\includegraphics[angle=-90,width=100mm]{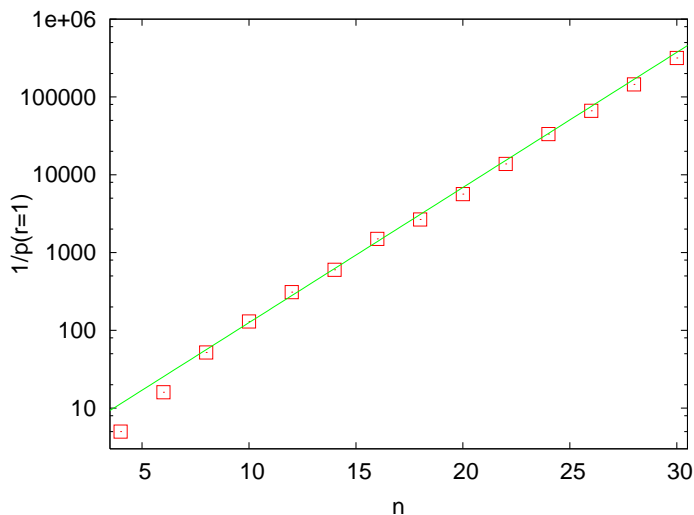}}
\caption{Frequency of 3-SAT formulas with exactly $r=1$ solution among random instances at $m/n=3$, i.e. the inverse probability to get such a formula. Each point is an average over $100$ instances.}
\label{fig:prob_ground}
\end{figure}
\begin{figure}[ht!]
\centerline{\includegraphics[angle=-90,width=100mm]{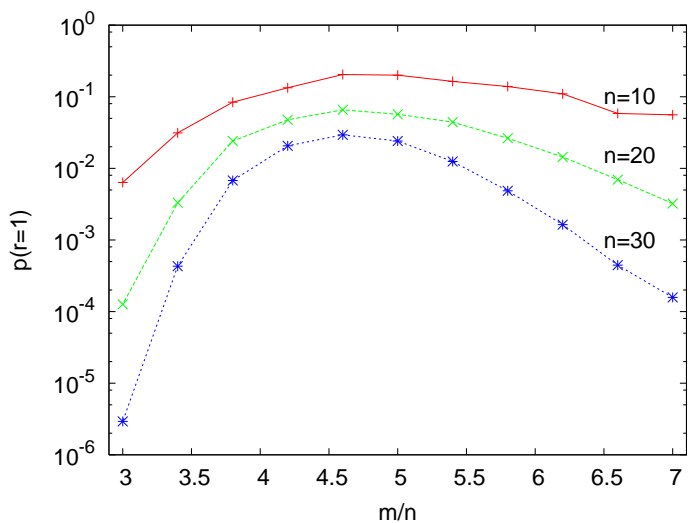}}
\caption{The dependence of probability $p(r=1)$ to find single-solution instance among random 3-SAT ensemble for three different $n$.}
\label{fig:prob_r1}
\end{figure}

Our ensemble therefore consists of random 3-SAT 
instances that have a single satisfying assignment ($r=1$). As the stress of this paper 
is to identify a new interesting ensemble of 3-SAT problems and not primarily 
to generate very large problems, we used a rather inefficient method to 
generate this ensemble. We obtained random 3-SAT instances with $r=1$ 
solution by simply filtering randomly generated formulas trough a 3-SAT 
solver keeping only those with $r=1$. The method is very inefficient because 
the expected number of solutions (below $\alpha_{\rm c}$) grows exponentially 
with $n$. Thereby, the probability to find an instance with only one solution 
among random 3-SAT ensemble decays exponentially with $n$. This can be seen in Figure~\ref{fig:prob_ground}. This rather technical 
issue of inefficient generation limited us to problems 
of relatively small size $n\le 40$. For instance, to generate $1000$ 
random 3-SAT instances with $n=40$, $r=1$ and $m/n=3$ we had to solve 
approximately $10^{10}$ randomly generated problems. Still, 
we think that it is conceivable to generate single-solution problems with constant $\alpha$ in a more efficient way, e.g. by converting one of the many known 
NPC problems~\cite{Garey:79} to 3-SAT with one solution. The probability to find single-solution formulas among random 3-SAT instances changes with $m/n$ and attains its maximum at the location of the transition point for random 3-SAT, see Figure~\ref{fig:prob_r1}.

\subsection{Algorithms used}
We will test several algorithms for solving 3-SAT problem. Generally there 
are two classes of algorithms : (i) a complete ones that determine 
satisfiability or unsatisfiability of a given formula. They terminate after 
a finite number of steps, either by finding a solution or proving 
unsatisfiability. (ii) incomplete algorithms which can only find a solution 
but can not prove unsatisfiability. In principle they have no terminating 
condition and can therefore be used only on satisfiable formulas on which 
though they can perform better than complete algorithms. Typically they employ some 
sort of a localized random search. 

The most popular complete method is the DPLL algorithm~\cite{DPLL:62}, 
sometimes called also just DP algorithm because it is based on an earlier 
algorithm by Davis and Puttnam~\cite{DP:60}. DPLL is a backtracking 
depth-first algorithm. It assigns truth values to variables and simplifies 
the formula. The simplification of formula $\phi$ when assigning TRUE 
value to a single literal $v$ consists of deleting clauses that are 
satisfied by the truth assignment, i.e. contain literal $v$, and deleting 
all literals contradicting the assignment in other clauses, i.e. all 
occurrences of $\lnot v$. The algorithm therefore descends along the 
state tree by recursive calls until it either finds a solution or 
encounters a contradiction, i.e. an empty clause occurs. In the later case 
it backtracks by changing a previously made assignment. The 
number of recursive calls of DPLL procedure is a good measure of 
running time. There are different variants of DPLL algorithm depending 
on the variable-selection rule, i.e. on the heuristics how we choose the 
next variable whose value we assign. In the algorithm we use we pick the 
first variable in the first unsatisfied clause. 
\begin{figure}[ht!]
\centerline{\includegraphics[width=95mm]{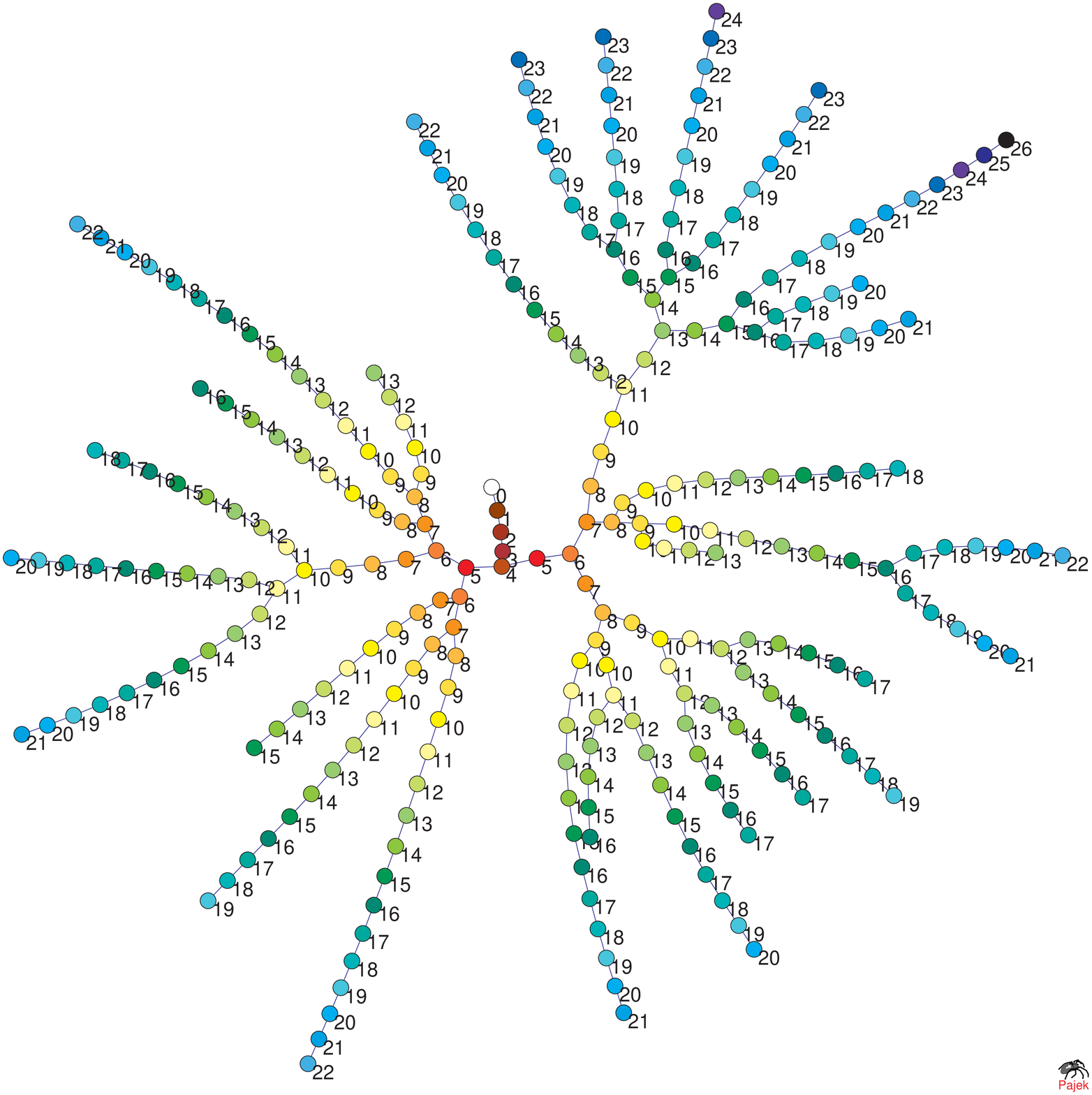}}
\caption{Search tree of the DPLL algorithm for 3-SAT instance with $n=26$ and $m/n=3$ having one solution, $r=1$. Color of vertices and numbers denote the number of assigned variables.}
\label{fig:sat26_3_depth}
\end{figure}
To illustrate DPLL algorithm we can plot a search tree as shown in 
Figure~\ref{fig:sat26_3_depth} for an instance with 
$n=26$ variables and $m/n=3$, having exactly one solution, $r=1$. 
Each vertex denotes a state with a certain number of assigned variables, 
the number of which is printed next to a vertex. The algorithm therefore 
starts from the white vertex with the number $0$ and ends at the black 
vertex with the number $26$, denoting a state which is a solution of 3-SAT 
formula. The number of DPLL calls needed to find the solution, equal to the
number of vertices, was $278$ in this case. The search tree has been plotted 
using the network analysis program ``Pajek''~\cite{pajek}.

Nowadays there are of course many modern algorithms that are much faster 
than the original DPLL one although most of them is still based on DPLL idea. Each year a competition for SAT solvers is organized comparing 
their performance on artificial and industrial SAT problems~\cite{SAT:05}. 
In addition to simple DPLL we have tested also one of those, namely SATO~\footnote{We used SATO v. 4.1.} by H.~Zhang~\cite{Zhang:97}. SATO is based on 
DPLL but uses different splitting rule and also uses 
``intelligent backjumping'', meaning that it must not backtrack step by 
step but can jump over several steps. In all numerical experiments the results
for SATO were qualitatively the same as for DPLL.

For satisfiable problems the so-called stochastic local search algorithms 
can be more effective than complete algorithms. We have used 
GWSAT~\cite{Selman:94} as our main stochastic algorithm. It is based on
GSAT~\cite{Selman:92,Selman:93}, and is one of the most widely studied 
incomplete methods. At the beginning of the algorithm we randomly draw a 
state, i.e. choose a random assignment of variables. Then at each step we 
change the truth value of one variable (such a step is also called a flip). 
For local search methods we need a cost function that will measure how good 
different flips are, so that we can choose the best one at each step. In 
GWSAT we choose with probability $1-p$ the variable to flip as the one which 
leads to the state with the largest number of satisfied clauses (GSAT step), 
and with probability $p$ a random variable from a randomly chosen unsatisfied clause. 
This is repeated until a solution is found. A good measure of running time is 
the number of flips made 
until a solution is found. In addition to GWSAT we also tested Walksat~\cite{Selman:94} and Adaptive Novelty+~\cite{adnovelty:02}. For all stochastic local 
search algorithm implementations we used the Ubcsat program~\cite{Ubcsat:04}.
Again, as for complete methods, the results for more advanced Walksat and Adaptive Novelty+ algorithms were qualitatively the same as for GWSAT. For a 
detailed comparison of different local search algorithms see, e.g.~\cite{Hoos:00}

In the next two subsections we will present the main results of the paper, 
the analysis of the running time of various 3-SAT solving algorithms for an 
ensemble of random 3-SAT instances with $1$ solution. First we will show the dependence of running time on $\alpha$ at constant $n$ in order to demonstrate that there is no phase transition-like peak in the difficulty. 

\subsection{Running time at constant $n$}

\begin{figure}[ht!]
\centerline{\includegraphics[angle=-90,width=100mm]{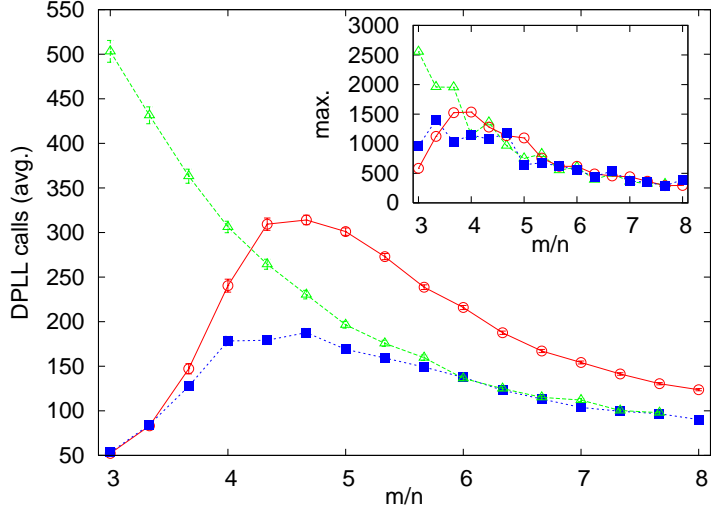}}
\caption{Running time for DPLL algorithm for random 3-SAT with $r=1$ 
solution (empty triangles), $r\ge 1$ solution (full squares) and arbitrary 
$r$ (empty circles). In the inset is shown the maximal running time 
(same three data sets) out of $1000$ instances used for each point. All is 
for $n=30$.}
\label{fig:DPLLphase}
\end{figure}

The data for DPLL are shown in Figure~\ref{fig:DPLLphase}. In addition to 
the curve for random instances with $r=1$ we also plot one for random 3-SAT 
ensemble (arbitrary $r$) and for random instances with $r \ge 1$ solutions. 
We can clearly see that the running time of instances with exactly one 
solution increases with decreasing $m/n$ and gets in fact larger for 
sufficiently small clause density than for instances around the phase 
transition point for random 3-SAT. The same quantitative results are obtained 
also for SATO. 

\begin{figure}[ht!]
\centerline{\includegraphics[angle=-90,width=100mm]{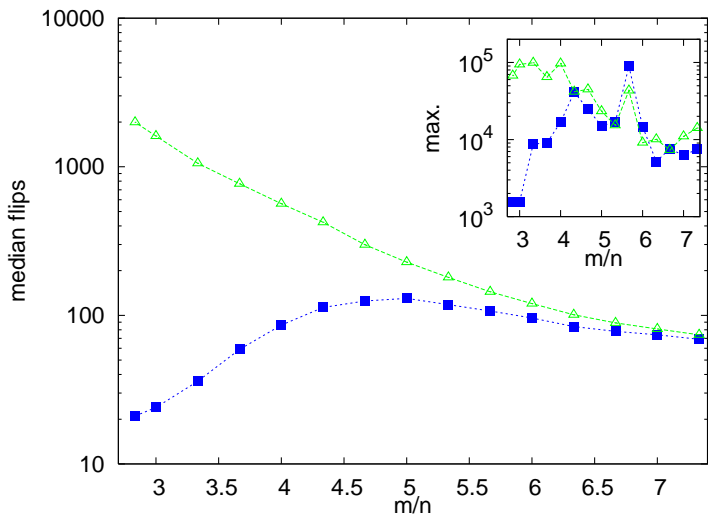}}
\caption{Median running time for GWSAT algorithm for random 3-SAT with $r=1$ 
solution (empty triangles) and $r\ge 1$ solution (full squares). In the 
inset is shown the maximal running time (same data sets) out of $1000$ 
instances used for each point. All is for $n=30$.}
\label{fig:GWSATphase}
\end{figure}
In Figure~\ref{fig:GWSATphase} we show the results for GWSAT algorithm. 
Again one can observe that the running time is larger for single-solution ensemble and that there is no peak in the difficulty. Similar figure is obtained also for Walksat and Adaptive Novelty+. Therefore, 
one can see that for an ensemble of single-solution random 3-SAT the 
difficulty monotonically increases with decreasing $m/n$. In comparison to random 3-SAT ensemble there is no peak in the difficulty for our ensemble. In view of that we will in
the next section study how the difficulty scales with $n$ for constant $m/n$. 
Because instances with smaller $m$ are harder, we will choose $m/n=3$. Note that by choosing even smaller $m/n$ one will get even harder 
instances, but then our inefficient generating method gets too slow. 

\subsection{Running time at constant $m/n$} 

Even though one can see in Figures~\ref{fig:DPLLphase} and~\ref{fig:GWSATphase} than 
instances from single-solution ensemble at $m/n=3$ are harder than the phase 
transition ones from random 3-SAT ensemble for the shown $n=30$, their 
difficulty could scale differently with $n$. The real question then is, what 
happens when we increase $n$.  
\begin{figure}[ht!]
\centerline{\includegraphics[angle=-90,width=100mm]{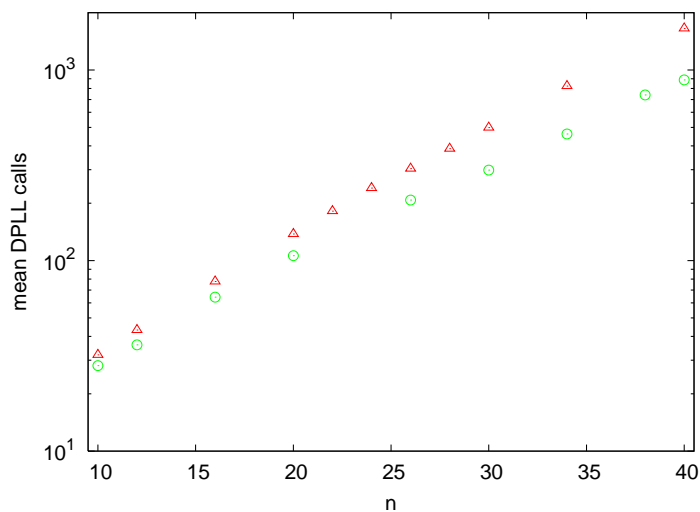}}
\caption{Scaling of the running time for DPLL algorithm. Triangles are for $r=1$ at $m/n=3$ and circles for arbitrary $r$ at $m/n=4.5$.}
\label{fig:skaldp}
\end{figure}
We will always compare results of a single-solution ensemble at $m/n=3$ and random 3-SAT ensemble at $m/n=4.5$ which is the approximate location of the transition point for random 3-SAT and small $n$ studied here. Each time we will average over 1000 3-SAT instances. In Figure~\ref{fig:skaldp} for DPLL algorithm one can see that the running time for single-solution instances increases faster than for random instances from the phase transition point. The difference in hardness therefore increases with increasing size.

\begin{figure}[th!]
\centerline{\includegraphics[angle=-90,width=100mm]{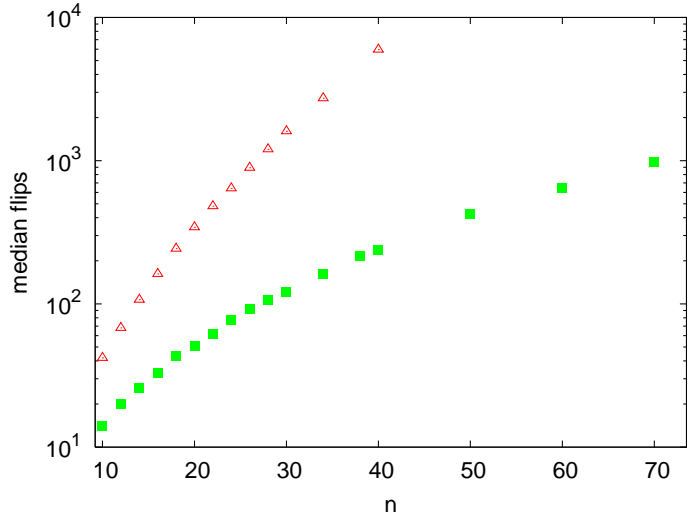}}
\caption{Scaling of the running time for GWSAT algorithm. Empty triangles are for $r=1$ at $m/n=3$ and full squares for $r\ge 1$ at $m/n=4.5$.}
\label{fig:skalgwsat}
\end{figure}
\begin{figure}[h!]
\centerline{\includegraphics[angle=-90,width=100mm]{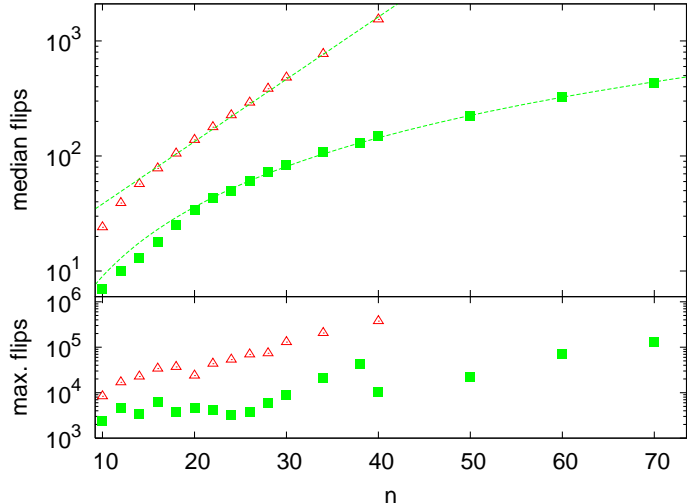}}
\caption{Scaling of the running time for Adaptive Novelty+. Empty triangles are for $r=1$ at $m/n=3$ and full squares for $r\ge 1$ at $m/n=4.5$. Dashed curves are $11\cdot 2^{0.18n}$ (top) and $0.09\cdot n^{2.0}$ (lower). In the lower plot we show the maximal times.}
\label{fig:skaladnovelty}
\end{figure}

Similar results can be observed also for GWSAT in Figure~\ref{fig:skalgwsat}. 
For stochastic search methods we average again over $1000$ instances and 
for each instance over $100$ runs of the algorithm. The difference again 
increases with increasing $n$, this time even faster as for DPLL. In Figure~\ref{fig:skaladnovelty} we show similar result also for Adaptive 
Novelty+. The number of necessary flips is smaller as for GWSAT but the 
overall behavior is qualitatively the same. Although the range of $n$ is 
relatively small we also plotted the best fitting exponential or power-law 
dependences. While for random 3-SAT instances from the phase transition one 
has a slow polynomial $\propto n^{2.0}$ growth of the running time, the 
increase is much faster, agreeing well with an exponential $\propto 2^{0.18 n}$, for single-solution ensemble. The coefficient of exponential growth $0.18$ is actually fairly large, meaning that short single-solution random instances get harder very quickly. For instance, 
for hard soluble 3-SAT instances reported in~\cite{Haixia:05} the increase 
of the number of flips for Walksat algorithm was approximately $\propto 2^{0.1 n}$ 
(with a large pre-factor). For our single-solution ensemble Walksat 
algorithm is slightly slower than Adaptive Novelty+. Interestingly, the same difference between the complexity scaling of two ensembles as here, namely polynomial vs. exponential, has been found also for quantum adiabatic algorithm~\cite{Farhi:01,Znidaric:05}. It might well be that the instances from the phase transition region of random 3-SAT ensemble are not that difficult and a polynomial average cost algorithm is possible. 

What is the explanation for the difficulty of single-solution random 3-SAT 
instances with small $m/n$? We will give some heuristic arguments why it might not be so unexpected that such 3-SAT instances are hard to solve. 

\begin{figure}[h!]
\centerline{\includegraphics[width=100mm]{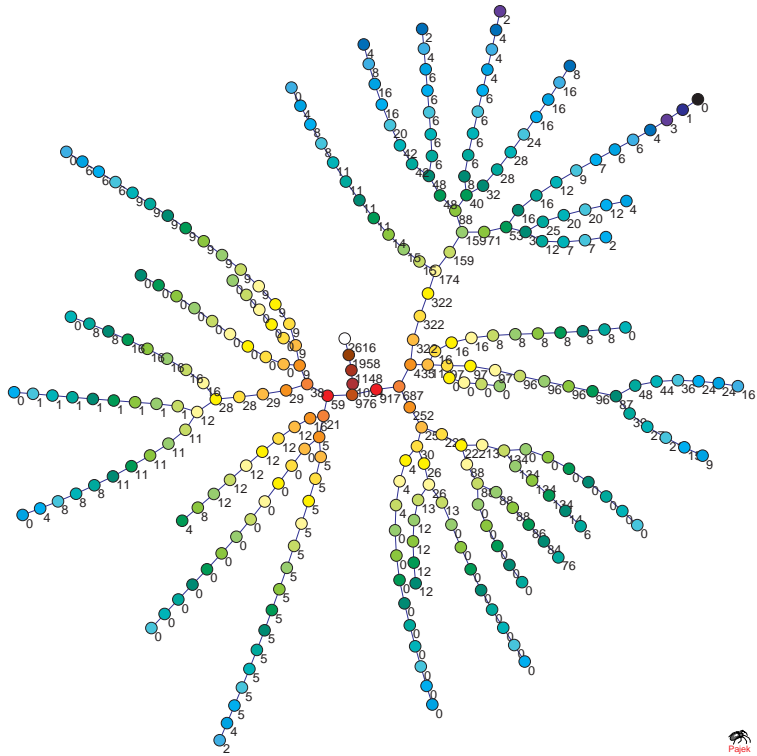}}
\caption{Search tree for DPLL and 3-SAT instance with $n=26$ and $m/n=3$ (the same instance as in Figure~\ref{fig:sat26_3_depth}). Numbers next to vertices denote the number of excited states in the sub-tree below a given vertex. The color of vertices denotes the depth in the tree as before.}
\label{fig:sat26_3_excit}
\end{figure}

For single-solution instances it turns out~\cite{Znidaric:05} that the number of assignments that violate only one clause, called the excited states, 
is very large. In fact the number of such excited states grows exponentially with $n$. Therefore, 3-SAT instances from single-solution random ensemble have 
only one solution and at the same time exponentially many assignments violating only one (or a few) clause. Any complete method, like DPLL for instance, must do exhaustive 
search in the state tree until a contradiction is encountered and the 
algorithm has to backtrack. If assignments made are such that we are 
descending towards an excited state which violates only one clause, a 
contradiction will occur only after we make an assignment of all three 
variables occurring in this clause. This can possibly occur very deep 
within the tree, causing large amounts of backtracking. Simply said, the 
excited states ``fake'' the algorithm into descending along the wrong 
branches. This can be seen in Figure~\ref{fig:sat26_3_excit}. This time 
the numbers next to vertices denote the number of excited states in the 
sub-tree below the vertex. We can see, that long branches are usually 
correlated with a large number of excited states. As a consequence, 
the search tree is large with many long branches. One can argue also 
differently: assuming a random truth assignment of the first assigned 
variable in a DPLL-like algorithm, one will with probability 
$\frac{1}{2}$ end up with an unsatisfiable problem. For this 
unsatisfiable problem a rigorous statement~\cite{Achlioptas:04} 
about the exponential complexity of DPLL even below the satisfiability 
border $\alpha_{\rm c}$ suggest an exponential complexity. Similarly, 
the running time is expected to be exponential also for incomplete stochastic 
local search methods. For such algorithms the exponential number of excited 
states will effectively shadow out the single real solution 
(searching for a needle in a haystack). To circumvent the exponential 
complexity the algorithm would have to efficiently distinguish between an 
exponentially many states violating only one clause and a single solution 
satisfying all clauses. This can not of course be excluded for some yet to be 
found smart choice of moves or a smart variable-selection rule in a 
DPLL-like algorithm, but it seems unlikely because there is simply very few 
information available which the algorithm could use in its heuristics. 
Remember that we are concentrating on underconstrained instances having as 
few clauses as possible. Of course, the explanation with excited states is probably only part of the story. It would be interesting to investigate the phase space structure of such instances in greater detail. 

\section{Conclusions}

We have identified and studied a new ensemble of 3-SAT instances, namely random 3-SAT formulae having a single satisfying assignment. Numerical experiments show that the properties of this ensemble are significantly different from those of random 3-SAT ensemble. The difficulty of single-solution instances monotonically increases with decreasing clause density, that is shorter formulas are generally harder. Therefore, this ensemble does not exhibit easy-hard-easy pattern of difficulty. Short single-solution instances (having e.g. $m/n=3$) are in fact much harder than problems from the phase transition region of random 3-SAT ensemble. It would be interesting to investigate the nature of their hardness more in detail. 3-SAT instances can be divided into ensembles according to the number of satisfying assignments they have. Here we studied only the ensemble having one satisfying assignment. An interesting question is, is the behavior of other ensembles similar, for instance that the difficulty of instances decays monotonically with $m/n$? If yes, the occurrence of the maximal difficulty for random 3-SAT at the phase transition can be viewed as being due to the changing of the probability to draw instances with fixed number of solutions. For single-solution instances studied in the present paper, the highest probability to find them among random 3-SAT instances occurs at the transition point and decays fast away from it. Becouse this decay is faster than the increase of their difficulty for small $m$, a maximum of difficulty occurs at the location of the maximum probability.

The author would like to thank the Alexander von Humboldt Foundation for its support.

\bibliography{hard3sat}
\bibliographystyle{elsart-num}

\end{document}